\documentclass[letterpaper, 10 pt, conference]{ieeeconf}  

\IEEEoverridecommandlockouts    
\usepackage{subcaption}
\usepackage{bm}
\usepackage{afterpage}
\usepackage{multirow}
\usepackage{graphics} 
\usepackage{epsfig} 
\usepackage{times} 
\usepackage{amsmath} 
\usepackage{amsfonts} 
\usepackage{color}
\usepackage{xcolor}
\usepackage{url}
\usepackage{hyperref}
\usepackage{booktabs}

\title{\LARGE \bf
Humanoid Whole-Body Locomotion on Narrow Terrain via Dynamic Balance and  Reinforcement Learning}

\author{Weiji Xie$^{1,2}$, Chenjia Bai$^{2*}$, Jiyuan Shi$^{2}$, Junkai Yang$^{1,2}$, Yunfei Ge$^{2}$, Weinan Zhang$^{1*}$, Xuelong Li$^{2}$
\thanks{$^{1}$Shanghai Jiao Tong University}
\thanks{$^{2}$Institute of Artificial Intelligence (TeleAI), China Telecom}
\thanks{*Correspondence to: Chenjia Bai (baicj@chinatelecom.cn), Weinan Zhang (wnzhang@sjtu.edu.cn).}
}

\newcommand{\ours}[0]{DBHL}

\DeclareMathAlphabet{\mathbbold}{U}{bbold}{m}{n}
\newcommand{\indicator}{\mathbbold{1}}

\newcommand{\authnote}[2][]{%
  \ifx&#1&%
    $\ll$\textsf{\footnotesize #2}$\gg$
  \else
    $\ll$\textsf{\footnotesize #1$\Vert$#2}$\gg$
  \fi
}

\begin{document}

\maketitle
\thispagestyle{empty}
\pagestyle{empty}

\begin{figure*}[!t]
    \centering
    \includegraphics[width=\textwidth]{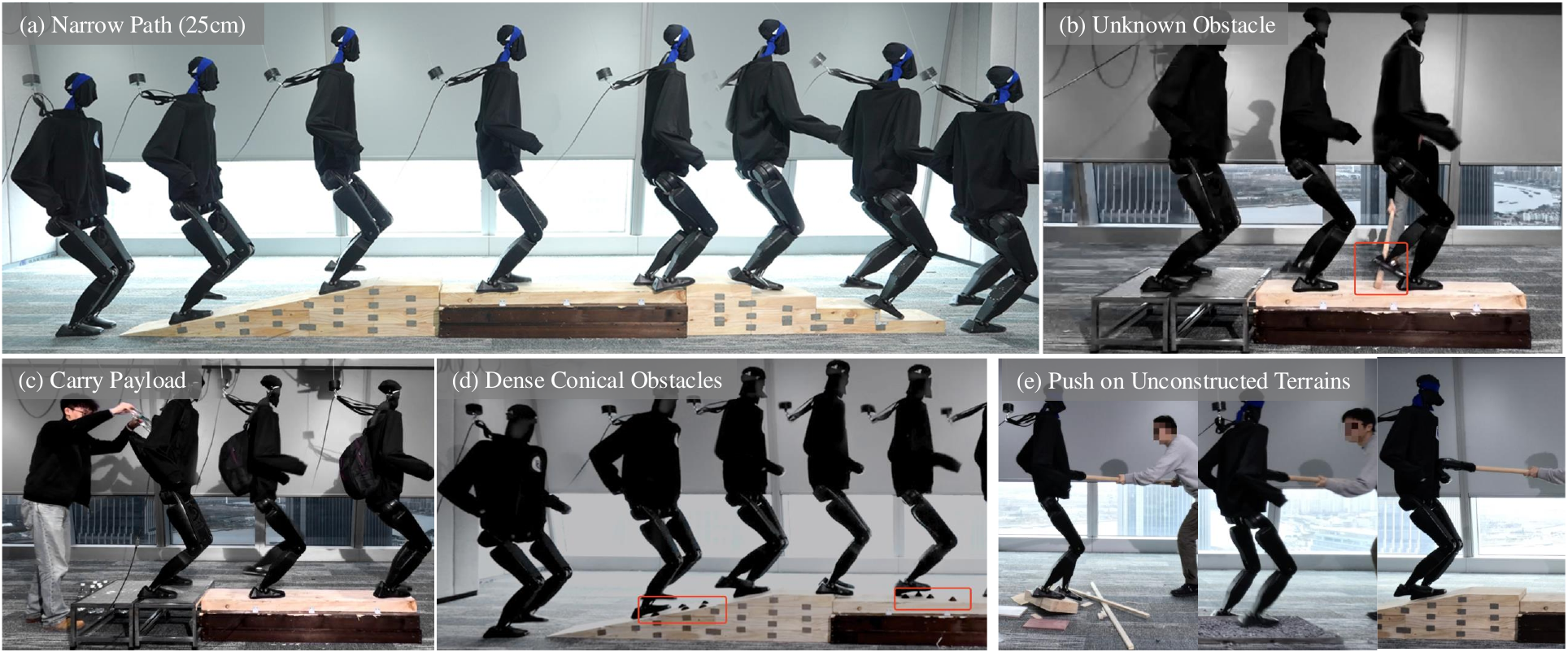}
    \caption{\textbf{The locomotion capabilities of full-sized Humanoid without vision or LiDAR sensors}. (a) \emph{Narrow Path (25cm)}: The humanoid traverses a narrow pathway, including slopes and stairs, demonstrating precise foot placement and dynamic balance. (b) \emph{Unknown Obstacle}: The humanoid robot showcases its dynamic balance control by swiftly adapting to the moving stick's attempts to trip it, maintaining stability even in this challenging scenario. (c) \emph{Carry Payload}: Our method can maintain stability while carrying loads, highlighting its robust control. (d) \emph{Dense Conical Obstacles}: The humanoid steps over a series of closely spaced cones, exhibiting agility and coordination. (e) \emph{External Pushes}: The system responds to external forces applied during locomotion over uneven terrain, proving its resilience against disturbances. Each scenario underscores the DBHL's versatility and effectiveness in handling complex conditions.}
    \label{fig:demo}
    \vspace{-1em}
\end{figure*}

\begin{abstract}

Humans possess delicate dynamic balance mechanisms that enable them to maintain stability across diverse terrains and under extreme conditions. However, despite significant advances recently, existing locomotion algorithms for humanoid robots are still struggle to traverse extreme environments, especially in cases that lack external perception (e.g., vision or LiDAR). This is because current methods often rely on gait-based or perception-condition rewards, lacking effective mechanisms to handle unobservable obstacles and sudden balance loss. 
To address this challenge, we propose a novel whole-body locomotion algorithm based on dynamic balance and Reinforcement Learning (RL) that enables humanoid robots to traverse extreme terrains, particularly narrow pathways and unexpected obstacles, using only proprioception. 
Specifically, we introduce a dynamic balance mechanism by leveraging a novel Zero Moment Point (ZMP)-driven reward and task-driven rewards in a whole-body actor-critic framework, aiming to achieve coordinated actions of the upper and lower limbs for robust locomotion. 
Experiments conducted on a full-sized Unitree H1-2 robot verify the ability of our method to maintain balance on extremely narrow terrains and under external disturbances, demonstrating its effectiveness in enhancing the robot's adaptability to complex environments. The videos are given at \href{https://whole-body-loco.github.io}{https://whole-body-loco.github.io}. 

\end{abstract}


\section{INTRODUCTION}

Recent advances in humanoid locomotion control have achieved significant progress, benefited from large-scale interaction and policy learning \cite{HumanGym1} in a Reinforcement Learning (RL) framework. These methods mainly include phase-based gait learning \cite{HumanGym2,HugWBC}, motor skill control \cite{decentralized,hilo}, and motion imitation \cite{AMP,embrace}. By leveraging  large-scale parallelized simulation \cite{Learning-to-walk} and policy optimization techniques \cite{PPO2}, current humanoid locomotion methods demonstrate well adaptation capabilities in conventional terrains. Despite these achievements, the locomotion ability of humanoid robots still lags far behind that of humans in terms of dynamic balance under extreme conditions. Especially, humans can quickly adjust their foot placements and centroids when faced with situations such as slipping or stepping off the edge, thus regaining stability. In contrast, current RL-based controllers lack such abilities primarily due to their reliance on periodic gait \cite{walk-this-way,HugWBC} or motion primitives \cite{HumanGym1,AMP}, which cannot achieve fast and diverse gait adjustment at critical moments of instability. We argue that a robust control policy should fully leverage the information on contact forces, the support polygon, and the centroid of the robot, which describes the fine-grained relationship between the robot and the support surface and is crucial for dynamic equilibrium.

Alternatively, classical biped locomotion research addresses this problem by considering two types of foot-ground contact during a walk cycle \cite{ZMP-1,ZMP-2}. Specifically, there is a double-support phase when the robot is supported on both feet, and a single-support phase when only one foot of the robot is in contact with the ground while the other is transitioning from the rear to the front position. In both cases, it is crucial to determine whether the contact can be maintained between the robot and the ground at a specific moment. Consequently, the concept of the Zero Moment Point (ZMP) is introduced to measure the influence of all forces acting on the robot that can be represented by a single force \cite{ZMP-3}. Specifically, the ZMP is defined as the point where the inertial and gravitational forces have no component along the horizontal axes. It has been demonstrated that if the ZMP lies within the support polygon of the foot and ground, the entire system is in dynamic balance. 
Inspired by this, we intergrate ZMP into learning-based humanoid whole body control, demonstrating significant improvements in the dynamic stability of humanoid robots when navigating complex terrains and resisting external disturbances.

In this paper, we propose a novel RL framework for whole-body locomotion in extreme scenarios, named \emph{Dynamic Balanced Humanoid Locomotion (\textbf{DBHL})}. To enhance the locomotion policy's ability to traverse complex terrains, particularly narrow pathways and sudden obstacles, we extend the concept of ZMP to non-planar surfaces, thereby forming a line of ZMPs. We then design a reward function for the RL policy that encourages the ZMP coordinates to be close to the center of the humanoid's support polygon. This reward function is calculated using privileged information obtained from simulations, while the policy is learned solely based on proprioception via an asymmetric actor-critic framework. This design allows the policy to be deployed in real-world scenarios without relying on external perception. Within the RL framework, we train a whole-body control policy that leverages upper-body swings to assist dynamic balance. We introduce angular momentum regularization and multiplicative action noise to constrain undesired body rotation and action range. 
Finally, we integrate the ZMP-based reward with command-following and regularization rewards using a reward vectorization technique, where each reward term is associated with a specific value function, avoiding inaccurate value estimation of small items in an accumulated reward.

In experiments, we evaluate \ours{} under various challenging conditions, including walking on surfaces with unknown disturbances, stepping on stairs with different widths and heights, and traversing narrow slopes with varying widths and degrees. The results demonstrate that \ours{} exhibits significantly better stability compared to other mainstream methods. We also analyze the role of the ZMP-reward through comparison, highlighting its crucial role in maintaining dynamic balance. Additionally, we conducted a detailed ablation study on other design elements. Our real-world experiments using the full-sized Unitree H1-2 robot illustrate the robot's capability to traverse narrow terrains, handle disturbances such as pushes and trips, and showcase enhanced stability and adaptability in various extreme scenarios, as shown in Fig.~\ref{fig:demo}.

The key contributions are summarized as follows.
\begin{itemize}
    \item We intergrate ZMP into the RL-based humanoid control framework as a novel reward function, realizing dynamic balance in complex terrains.
    \item We construct a whole-body control framework with newly introduced techniques including reward vectorization, angular momentum regularization, and multiplicative action noise.
    \item We evaluate the proposed method via extensive experiments in both the simulation and the real-world using full-sized humanoid robots.
\end{itemize}




\section{RELATED WORK}

\subsection{Humanoid Locomotion}

Research on humanoid locomotion can be traced back to the 1970s \cite{Kato1974InformationPower}. The fundamental idea for a locomotion controller is to decompose it into planning and tracking modules, where the planning module is responsible for generating desired trajectories and the tracking module ensures that the robot follows these trajectories accurately \cite{grandia2023perceptive,meduri2023biconmp}. Methods such as Whole-Body Control (WBC) and Model Predictive Control (MPC) have achieved significant success in this domain \cite{li2023multi,sentis2006whole,sleiman2021unified}. However, these methods typically require precise modeling of the dynamics \cite{koenemann2015whole,schultz2009modeling}, which poses substantial challenges for complex robot structures. In recent years, learning-based algorithms have emerged as a promising alternative to legged locomotion \cite{HumanGym1,ha2024learning} with efficient parallel simulation \cite{Learning-to-walk}, significantly reducing the cost of interaction between the robot and the environment. Policies trained extensively in simulation environments can then be transferred to real robots \cite{he2024bridging}. In the field of quadrupedal locomotion, RL algorithms have demonstrated excellent performance in complex tasks such as complex-terrain walking \cite{levy2024learning,shi2024robust}, gait control \cite{han2024lifelike,huang2024diffuseloco}, and even parkour \cite{zhuang2023robot,cheng2024extreme}. 

For RL-based humanoid locomotion, things become more difficult due to their limited support areas and high center of gravity. Meanwhile, classical control algorithms are also limited by the inaccurate modeling of the complex dynamical system. Recent approaches have proposed using RL algorithms for phase-based gait learning \cite{HumanGym2,HugWBC}, motor skill control \cite{decentralized,hilo}, and motion imitation \cite{AMP,embrace}. However, these methods still lags far behind that of humans in terms of dynamic balance under complex terrain (e.g., narrow paths) and extreme conditions (e.g., sudden disturbance). In our work, we address this problem by measuring the relationship between the ZMP and the humanoid support polygon in an RL framework. 
We note that a concurrent work uses foothold rewards to pass through narrow areas, while it relies on a LiDAR-based elevation map for real-world deployment \cite{wang2024beamdojo}. In contrast, our method can traverse complex terrains only using proprioception without vision or LiDAR perception.

\subsection{Zero Moment Point}

ZMP became a crucial tool in classical humanoid locomotion a few decades ago \cite{ZMP-1}, which provides a framework for ensuring dynamic stability in bipedal robots. Formally, ZMP is defined as the point on the ground at which the net moment of the inertial forces and the gravity forces has no component along the horizontal axes \cite{ZMP-2}. In subsequent research, the ZMP concept has been instrumental in gait synthesis and has been integrated with advanced sensors to facilitate real-time balance adjustments \cite{ZIP-35}. In addition, it has inspired the exploration of innovative materials and foot designs to enhance the interaction of robots \cite{ZMP-book}. 
In our work, we extend the ZMP as a reward function to measure the relationship between the line of ZMPs and the support polygon, which enables the humanoid to maintain dynamic balance in complex terrains without relying on external perception. 

\section{METHOD}

\begin{figure}[t]
  \centering
  \includegraphics[width=0.5\textwidth]{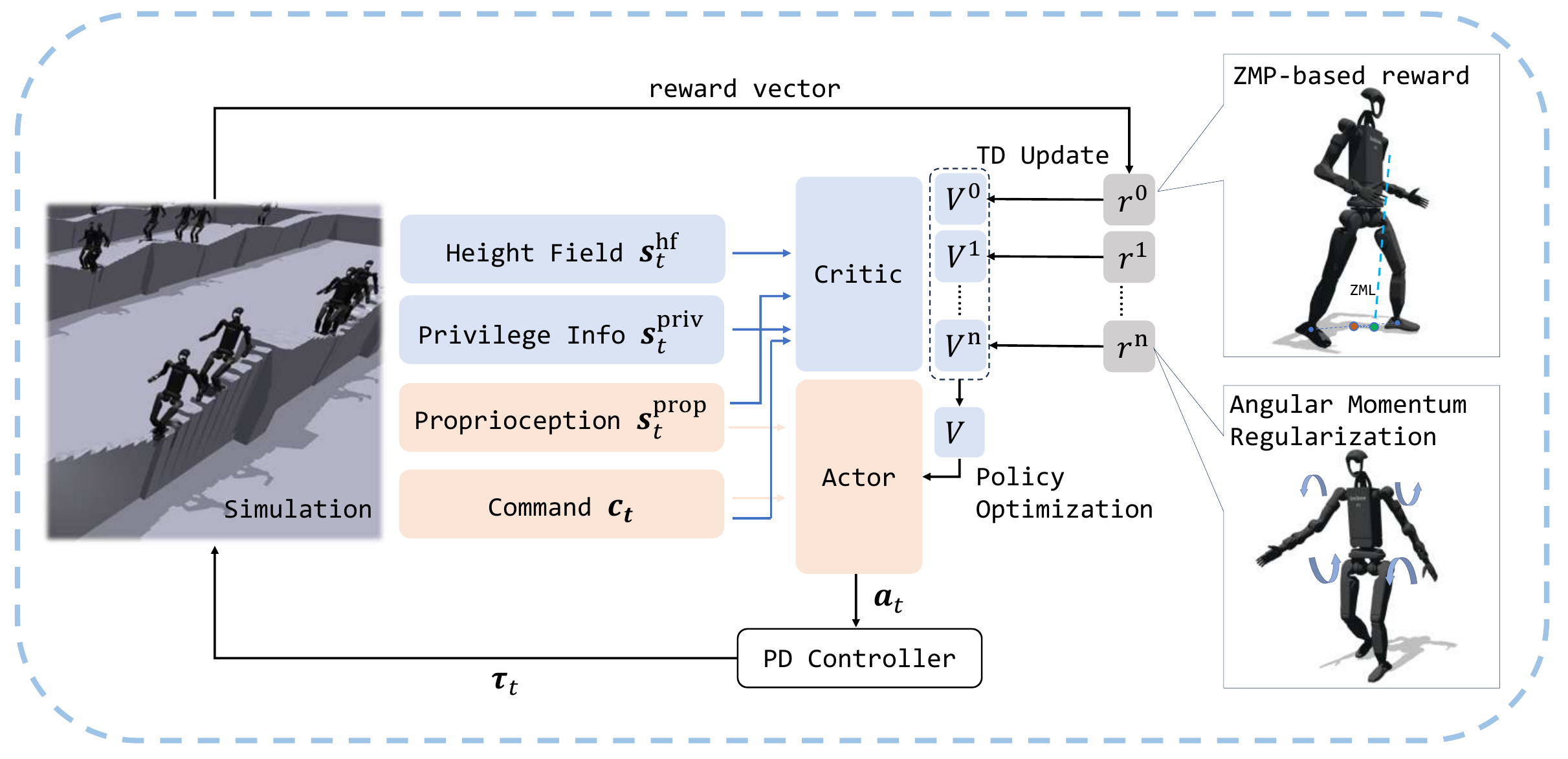}
  \caption{The overall training process of the proposed method.}
  \label{fig:method}
\vspace{-1em}
\end{figure}

In this section, we present our method for training an end-to-end RL policy that enables humanoid robots to traverse extreme terrains using only proprioceptive information. We formulate our problem as a goal-conditioned RL task, where the policy $\bm \pi$ is trained to follow the target velocity command. The action $\bm a_t$ represents the target joint positions, which are fed into the PD controller to actuate the robot's degrees of freedom. The agent's observation $\bm o_t$ comprises velocity command $\bm c_t$ and the history of proprioception information $\bm s^\text{prop}_t$. We employ the Proximal Policy Optimization (PPO) algorithm \cite{PPO1} with asymmetric actor-critic networks \cite{AAC} to maximize the cumulative discounted reward of the policy. The whole architecture is shown in Fig.~\ref{fig:method}.



\subsection{ZMP-based Dynamic Balance}

\begin{figure}[t]
  \centering
  \begin{subfigure}[c]{0.15\textwidth}
    \centering
    \includegraphics[width=\textwidth]{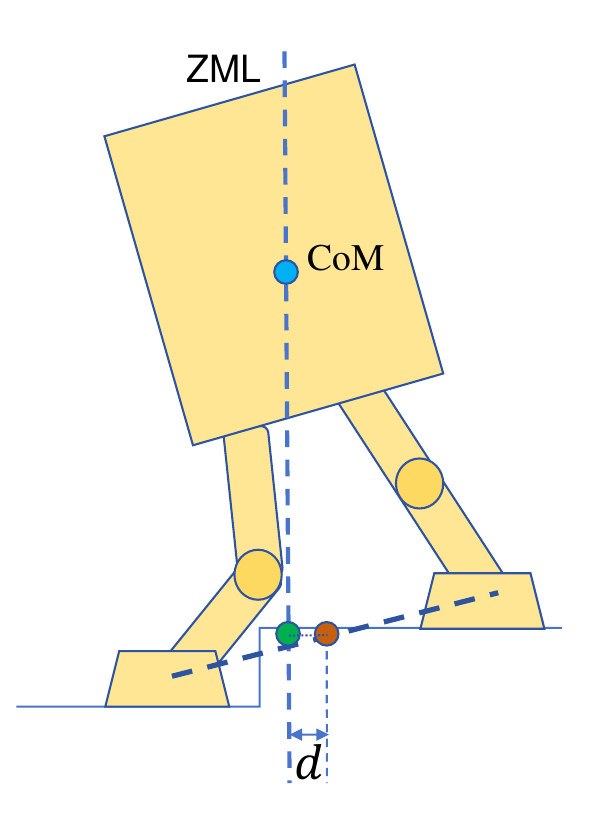}
    \caption{Standing still with two feet}
    \label{fig:ZML_static}
  \end{subfigure}
  \hfill
  \begin{subfigure}[c]{0.15\textwidth}
    \centering
    \includegraphics[width=\textwidth]{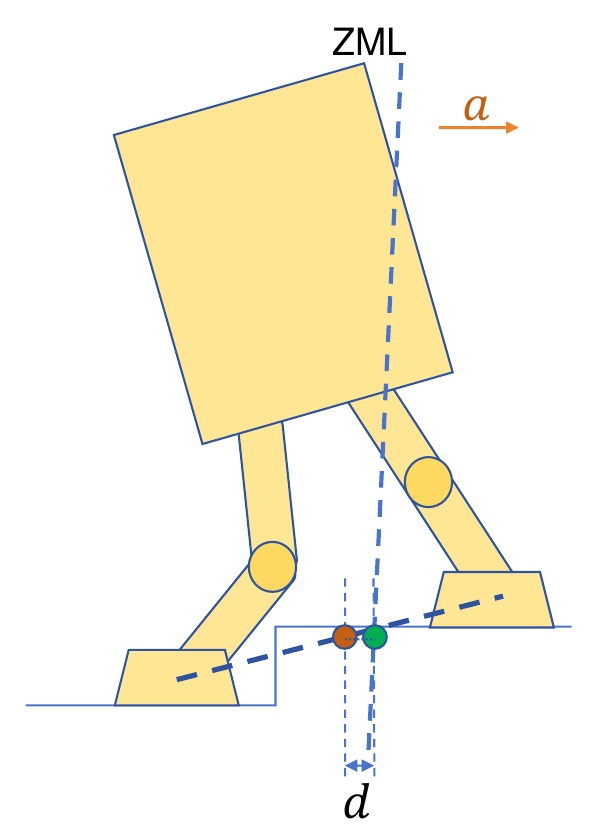}
    \caption{Leaning forward with two feet}
    \label{fig:ZML_dynamic}
  \end{subfigure}
  \hfill
  \begin{subfigure}[c]{0.15\textwidth}
    \centering
    \includegraphics[width=\textwidth]{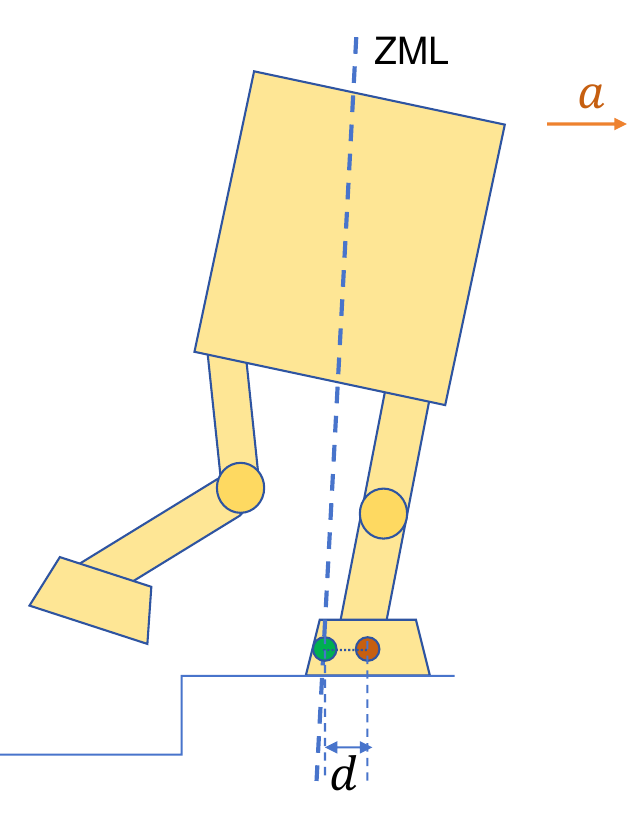}
    \caption{Stepping forward with one foot}
    \label{fig:ZML_single}
  \end{subfigure}
  \caption{Illustration of ZMP-based reward in different locomotion conditions. The brown dot represents the approximated center of the support polygon, $\bm p_\text{csp}$, and the green dot is the projection of point $\bm p_\text{csp}$ onto the ZML in the horizontal plane.}
  \label{fig:ZML}
  \vspace{-1em}
\end{figure}

To achieve dynamic balance, we design a ZMP-based reward and integrate it into the RL framework. 
The reward is calculated by the distance between a line of ZMPs, called Zero Moment Line (ZML) \cite{ZML}, and the center of the support polygon formed by the robot's feet, as shown in Fig.~\ref{fig:ZML}.


Assuming that the ground reaction force is the only external force applied to the robot system, the moment of the ground reaction force about the origin of the world frame is given by 
\begin{equation}
\bm \tau = \bm{p}_\text{zmp}\times \bm f+\bm{\tau}_p,
\label{eq:ground_tau}
\end{equation}
where $\bm{p}_\text{zmp}$ is the position of ZMP, $\bm{\tau}_p$ is the moment about the ZMP, and  $\bm f$ represents the ground reaction force. By Newton-Euler equations, we have the following relationship
\begin{equation}
\left\{ \begin{array}{l}
\dot{\bm{\mathcal P}} = M \bm g + \bm f \\
\dot{\bm {\mathcal L}} = \bm{p}_{\text{CoM}} \times M \bm g + \bm \tau,
\end{array} \right.
\end{equation}
where $\bm{\mathcal P}, \bm{\mathcal L}, M, \bm{p}_\text{CoM}$ represents the linear momentum, the angular momentum about the origin, the total mass, the center of mass (CoM) position of the system, respectively, and $\bm g = [0,0,-g]^\top$. By the definition of ZMP, we have
\begin{equation}
    \tau_{p,x}=\tau_{p,y}=0.
\end{equation}
Solving the above equations with respect to $\bm{p}_\text{zmp}$, we have
\begin{align}
p_{\text{zmp},x} = \frac{M  g p_{\text{CoM},x} + p_{\text{zmp},z} \dot{\mathcal P}_x - \dot{\mathcal L}_y}{Mg + \dot{\mathcal P}_z} \label{eq:px}
\\
p_{\text{zmp},y} = \frac{Mgp_{\text{CoM},y} + p_{\text{zmp},z} \dot{\mathcal P}_y + \dot{\mathcal{L}}_x}{Mg + \dot{\mathcal P}_z}.
\label{eq:py}
\end{align}
According to Eq.~\eqref{eq:px} and Eq.~\eqref{eq:py}, ZMP at different heights $p_{\text{CoM},z}$ lies at different locations, whose trajectory forms a line of ZMPs, called ZML \cite{ZML}. 




The condition for stable locomotion is that the ZML must cross the support polygon.  
In planar surfaces, the support polygon is determined by computing the convex hull of all contact points for scenarios with multiple contact points. However, the calculation becomes intricate when dealing with irregular contact surfaces, such as slopes or uneven terrain, where contact points need to be projected onto a virtual plane \cite{ZMP-multicontact}. 
In \ours, we bypass explicit computation of the support polygon through reward design. In this way, the problem of dynamic balance is transformed into the design of a metric, which encourages the ZML to be close to the center of the support polygon.

Specifically, the geometric \underline{c}enter of \underline{s}upport \underline{p}olygon (denoted as $\bm{p}_\text{csp}$) is approximated by the center of supporting feet, as
\begin{equation}
\begin{aligned}
\bm{p}_\text{csp}=\frac{\bm{p}_\text{left-foot} \cdot (c_\text{left-foot} \!+\! \epsilon) \!+\! \bm{p}_\text{right-foot} \cdot (c_\text{right-foot} \!+\! \epsilon)}{c_\text{left-foot} + c_\text{right-foot} + 2\epsilon},
\label{eq:csp}
\end{aligned}
\end{equation}
where $\bm{p}_\text{left-foot}$ and $\bm{p}_\text{right-foot}$ are the center of the left and right foot, respectively; $c_\text{left-foot}=
\indicator[\|\bm{f}_\text{left-foot}\|_2 \!>\! 0]$ and $c_\text{right-foot}=
\indicator[\|\bm{f}_\text{right-foot}\|_2 \!>\! 0]$ are indicator functions to determine whether the foot is in contact with the ground.
According to Eq.~\eqref{eq:csp}, if a humanoid robot is supported by two feet, $\bm{p}_\text{csp}$ is approximately equal to the center of two feet. And if a humanoid robot is supported by one foot, $\bm{p}_\text{csp}$ is approximately equal to either
$\bm{p}_\text{left-foot}$ or $\bm{p}_\text{right-foot}$.

Based on the simplified $\bm{p}_\text{csp}$, we design a computationally tractable reward function, denoted as $r_\text{zmp}$, with the horizontal distance between $\bm{p}_\text{csp}$ 
and the ZML, called ZMP-distance, as 
\begin{equation}
r_\text{zmp} = \exp\left(-{\Vert \bm{p}_\text{csp} - \text{Proj}_\text{ZML}(\bm{p}_{\text{csp}})  \Vert_2}/{0.05}\right),
\label{eq:rzmp}
\end{equation}
where $\text{Proj}_{\text{ZML}}(\bm{p}_{\text{csp}})$ is a projection function that projects $\bm{p}_{\text{csp}}$ to the ZML in the horizontal plane. Intuitively, dynamic stability is guaranteed when $\bm{p}_\text{csp}$ is close to the ZML, and a smaller distance indicates better stability, as shown in Fig.~\ref{fig:ZML}. 

\subsection{Whole-Body Locomotion}

To achieve coordinated whole-body motion and enhance dynamic stability, we propose a whole-body locomotion framework that leverages upper-body swings to assist dynamic balance. To achieve this, we incorporate two key techniques: angular momentum regularization and multiplicative action noise injection.

\subsubsection{Angular Momentum Regularization} It is introduced to minimize the undesired rotational motion during locomotion, thus improving coordination and improving resistance to external disturbances. As extensively discussed in prior work \cite{angularmomentum,highspeedlocomotion}, the motion of swinging legs generates significant angular momentum, which can disrupt whole-body motion. This effect can be mitigated through the proper use of upper limbs to counterbalance the angular momentum. We introduce a regularization reward $r_\text{AM}$ based on the $L_2$ norm of the total angular momentum about the robot base position, $\bm{\mathcal L}_
\text{base}$, which is defined as 
\begin{equation}
r_\text{AM}=\exp\left(-{\Vert \bm{\mathcal L_\text{base}} \Vert_2}/{5}\right),
\label{eq:ram}
\end{equation}

\begin{equation}
\bm{\mathcal L}_
\text{base}=\sum\limits_{i=1}^n \bm{p}^\text{base}_i\times (m_i \bm{v}^\text{base}_i)+\bm{I}_i \bm{\omega}^\text{base}_i,
\label{eq:am}
\end{equation}
where $m_i$ represents the mass of $i$-th link, $\bm{p}^\text{base}_i$, $\bm v^\text{base}_i$, $\bm \omega^\text{base} _i$ are the CoM position, linear velocity, and angular velocity of the $i$-th link relative to the base position, respectively, and $\bm I_i$ is the inertia tensor of the $i$-th link about its CoM.
 
\subsubsection{Multiplicative Action Noise Injection}\label{sec:acn} It is employed to constrain the range of motion of the upper body joints to enhance the robustness of the policy. If the policy is not restricted, the locomotion policy can often result in unstable and large-angle upper limb movements. Such a technique is implemented by modifying the input to the PD controller, where the nominal action \(a_t\) is perturbed by a multiplicative noise term, as 
\begin{equation}
\bm{a}_t^\prime=\bm{a}_t(1+\sigma_\text{AN} \bm{\epsilon}_t),
\label{eq:acn}
\end{equation}
where $\bm{\epsilon}_t\sim N(0,\mathbf I)$. By applying large perturbations to actions with greater magnitudes, the policy is encouraged to favor small actions for the upper body joints and to promote cautious behavior when significant movements are required.

\subsection{Policy Learning Details}

\subsubsection{Task and Command}

We developed three types of narrow terrain in policy training, including flat, stairs, and slopes. The robot is trained in a mixing of these narrow terrains, each containing 30\% with an additional 10\% planar terrain. Meanwhile, a terrain-curriculum mechanism introduced in \cite{Learning-to-walk} is used in training with 20 difficulty levels, where the path width gradually decreases from 1.0 to 0.2m, the slope gradient increases from 0 to 0.3, and the step height increases from 0 to 0.12m. 
For each episode, the linear velocity command is uniformly sampled in the range $\hat v_{x,t}\in[-0.5 \text{m/s},1.0 \text{m/s}]$ and $\hat v_{y,t}\in[-0.2 \text{m/s},0.2 \text{m/s}]$, while the yaw velocity command is determined by $\hat \omega_{\text{yaw},t}=\text{clip}(0.5*\Delta\theta_\text{yaw},-1 \text{rad/s},1 \text{rad/s})$ where $\Delta\theta_\text{yaw}$ is the horizontal angle between positive X-axis direction and the orientation of the robot.

\subsubsection{Asymmetric Actor-Critic Framework}

The observation for the actor network 
$\bm o_t\in \mathbb R^{279}$ comprises velocity command $\bm c_t=[\hat v_{x,t}, \hat v_{y,t},\hat \omega_{\text{yaw},t}]$ and the robot's proprioception $\bm s^\text{prop}_t=[\bm q_{t-3:t},\dot {\bm q}_{t-3:t}, \bm \omega_{t-3:t},\bm g_{t-3:t},\bm a_{t-4:t-1}]$ with 4-step history of joint position $\bm q_t\in \mathbb R^{21}$, joint velocity $\dot {\bm q}_t\in \mathbb R^{21}$, base angular velocity $\bm \omega_t\in \mathbb R^{3}$, base projected gravity $\bm g_t\in \mathbb R^{3}$ and last action $\bm a_{t-1}\in \mathbb{R}^{21}$. The observation for the critic network $\bm s_t$ includes actor observation $\bm o_t$, privileged information $\bm s^\text{priv}_t\in \mathbb R^{70}$ and surrounding height field $\bm s^\text{hf}_t\in \mathbb R^{187}$. The privileged observation $\bm s^\text{priv}_t$ contain the linear velocity, the base height, feet contact indicator, randomized PD parameter and randomized link mass. The height field $\bm s^\text{hf}_t$ is sampled from a $1.6 \text{m} \times 1.0\text{m}$ area around the robot, with a point spacing of $0.1$m. This framework leverages privileged information to enhance value function estimation for policy guidance, while restricting the actor to a local state ensures the policy's transferability to real-world environments. 

\subsubsection{Vectorization of Reward and Value}
To facilitate the learning of the value function with different reward functions, we introduce the vectorization of the reward and value function. Instead of aggregating all reward terms into a single scalar and learning a single value function, we combine the different rewards as a vector and learn the corresponding value functions via Temporal-Difference (TD) learning independently. Then, we obtain a set of value functions, each associated with a specific TD target. To achieve this, the value function is implemented by a neural network with multiple output heads. Then, all value functions are aggregated in computing the action advantage function. This method addresses a key limitation in traditional approaches, where summing all rewards makes it difficult for the value function to capture changes of reward terms with relatively small magnitudes. Specifically, we have
\begin{equation}
V^\text{total}(\bm s_t)=\sum\nolimits_{i=1}^{\text{\#Reward}}V^i(\bm s_t),
\end{equation}
and the loss function to train a value function is given by
\begin{equation}
{L}_\text{value}=\sum\nolimits_{i=1}^{\text{\#Reward}}\mathbb E\Big[\big\Vert  r^i_t+\gamma V^{i}(\bm s_{t+1})-V^{i}(\bm s_t) \big\Vert^2_2\Big].
\end{equation}
The reward functions of DBHL are given in Table~\ref{tab:rew}.

\begin{table}[!ht]
    \centering
    \caption{Reward functions of DBHL. The feet contact reward encourages single contact between the feet and the ground. The feet edge distance reward encourages the feet to stay far from the ground edge. The action closeness reward encourages the action to be close to the current DOF position.}
    \label{tab:rew}
    \begin{tabular}{lp{6cm}}  
        \toprule
        Group & Reward Function \\ 
        \midrule
        Task & Linear Velocity Tracking, Angular Velocity Tracking, Low Speed Penalty \\ 
        \midrule
        Gait & ZMP, Feet Air Time, Feet Contact, Feet Separation, Feet Slippage, Feet Height, Base Height, Feet Edge Distance \\ 
        \midrule
        Regularization & Angular Momentum, Orientation, Base Acceleration, Action Smoothness, Action Closeness, Torque, DOF Velocity, DOF Position Limit, Collision \\ 
        \bottomrule
    \end{tabular}
\end{table}

\subsubsection{Symmetry Regularization} 

Inspired by \cite{Symmetry}, we introduce a symmetry loss term to enhance sample efficiency and promote more harmonious gaits. This loss leverages the symmetry of the robot's motion with respect to the \(x\)-\(z\) plane, defined as
\begin{equation}
 L_\text{symm}=\mathbb E \big[ \Vert \bm V(G(\bm s_t))-\bm V(\bm s_t)\Vert^2_2+\Vert\bm \pi(G(\bm o_t))-G(\bm \pi(\bm o_t))\Vert^2_2 \big],
\label{eq:lsymm}
\end{equation}
where $G$ denotes the reflection operator across the \(x\)-\(z\) plane.

\subsubsection{Domain Randomization}

\begin{table}[!ht]
    \centering
    \caption{Domain randomization settings.}
    \begin{tabular}{lp{6cm}}
    \toprule
        Term & Value \\ \midrule
        External Push & $\Delta T\sim \text{Exp}(6)\text{ s}$, $\Delta \bm v_{xy}\sim \mathcal U(-0.6,0.6)\text{ m/s}$, \\&$\Delta \bm\omega\sim \mathcal U(-0.8,0.8 )\text{ rad/s}$ \\ 
        Action Delay & $\mathcal U(4,20)$ ms \\ 
        P Gain & $\mathcal U(0.8,1.2)\times$ default \\ 
        D Gain &  $\mathcal U(0.8,1.2)\times$ default \\ 
        Friction & $\mathcal U(0.1,2)$ \\ 
        Link Mass & $
        \mathcal U(0.8,1.2)\times$ default \\ 
        Load Mass & $\mathcal U (-1,3)$ kg \\ 
        Base CoM Offset & $\mathcal U(-0.1,0.1)$ m \\ 
        Torque RFI & $\Delta\bm\tau\sim\mathcal U(-0.1,0.1 )\times\tau_\text{rfi}\times$ torque limit N$\cdot$m, \\&$\tau_\text{rfi}\sim \mathcal U(0.5,1.5)$ for each episode\\ 
        Action Noise & $\sigma_\text{AN}=0.03$, see Sec.~\ref{sec:acn}. \\ 
        \bottomrule
    \end{tabular}
    \label{tab:dr}
\end{table}

To enable zero-shot sim-to-real transfer, we randomize the physical parameters of the simulated environment and humanoids, listed in Table~\ref{tab:dr}.


\begin{figure}[t]
  \centering
  \begin{subfigure}[c]{0.239\textwidth}
    \centering
    \includegraphics[width=\textwidth]{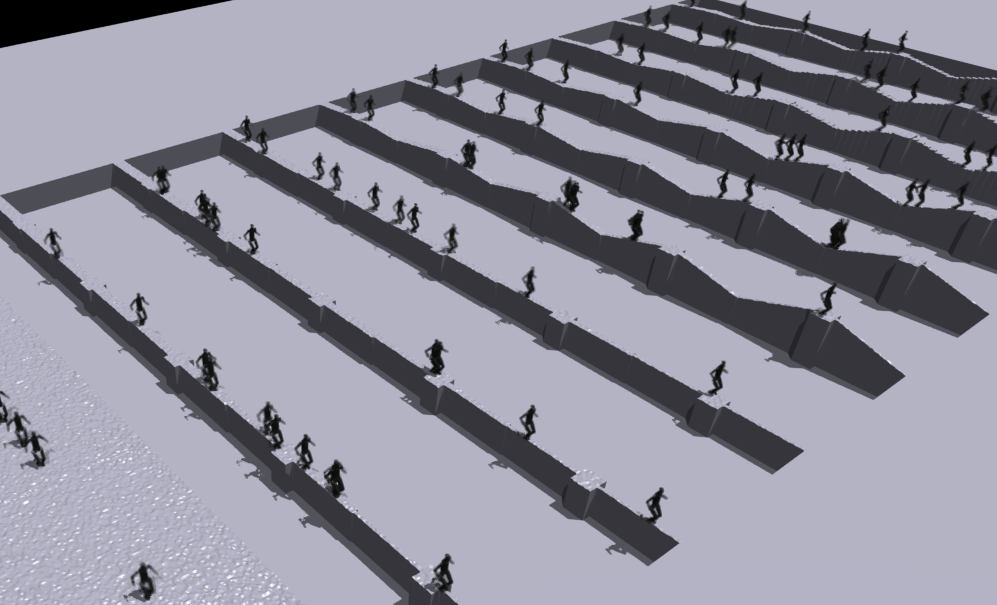}
  \end{subfigure}
  \hfill
  \begin{subfigure}[c]{0.239\textwidth}
    \centering
    \includegraphics[width=\textwidth]{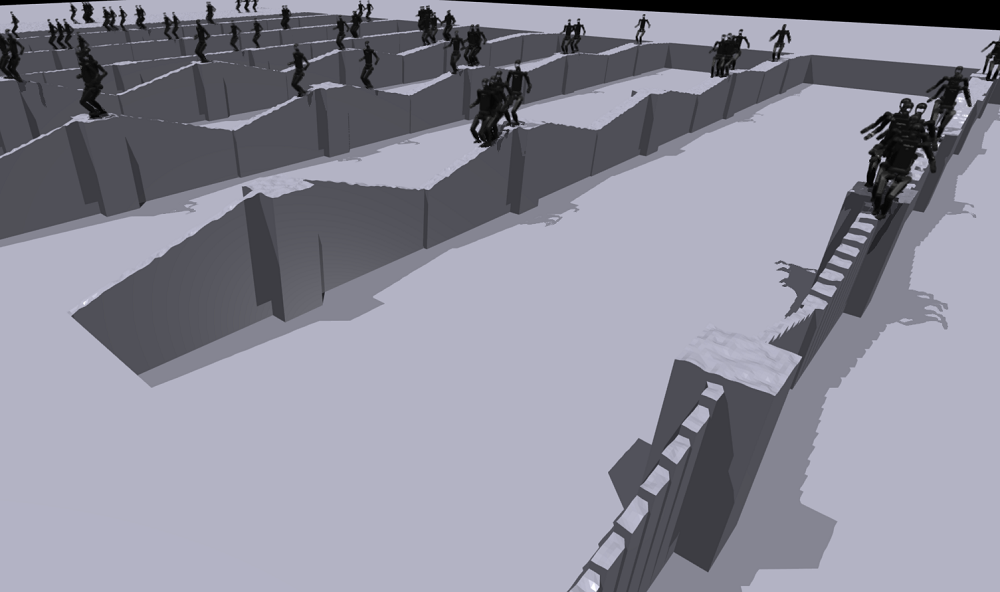}
  \end{subfigure}
  \\ \vspace{0.5em}
  \begin{subfigure}[c]{0.239\textwidth}
    \centering
    \includegraphics[width=\textwidth]{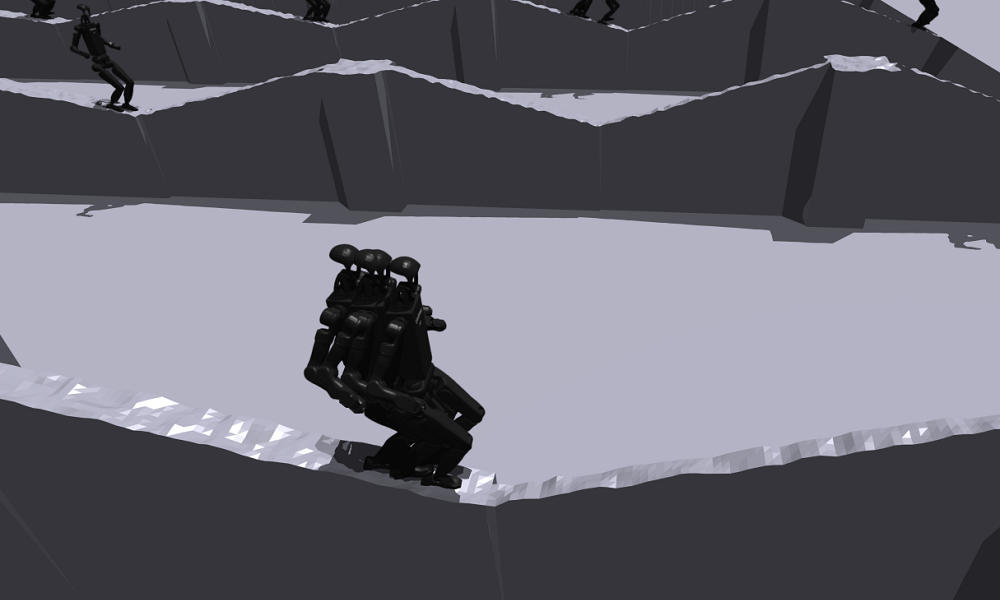}
  \end{subfigure}
  \hfill
  \begin{subfigure}[c]{0.239\textwidth}
    \centering
    \includegraphics[width=\textwidth]{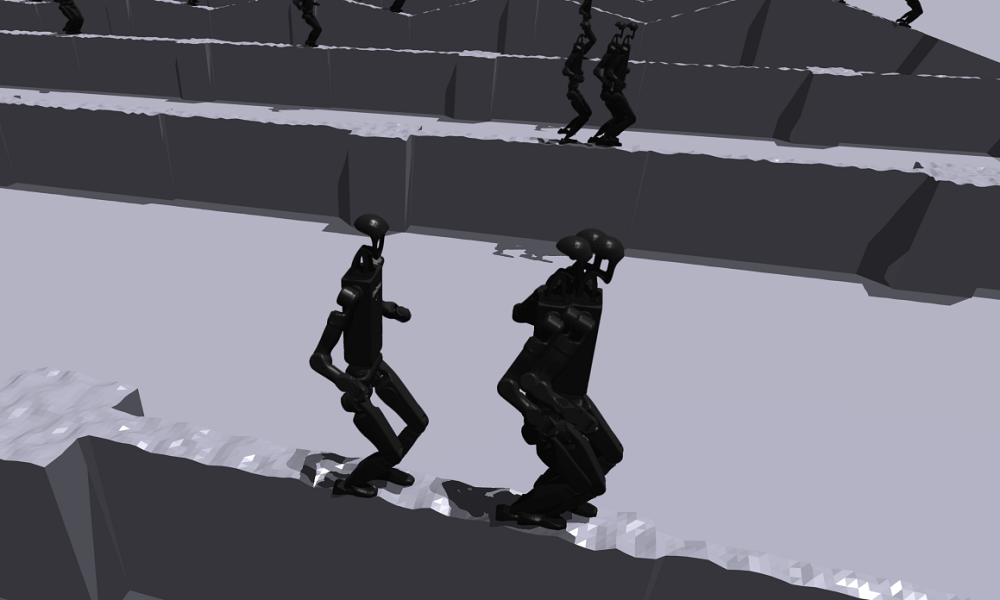}
  \end{subfigure}
  \caption{Visualization of the various training terrains of our method in Isaac Gym.}
  \label{fig:terr}
\vspace{-1.5em}
\end{figure}

\section{EXPERIMENT}

In this section, we present extensive experiments to evaluate the proposed method. Our experiments aim to address the following key questions: 
\begin{itemize}
    \item Q1: Can \ours{} outperform other methods in various extreme terrains?
    \item Q2: How does the ZMP-based reward contribute to dynamic balance?
    \item Q3: How do design choices (i.e., reward vectorization and action noise) influence training performance?
    \item Q4: Can \ours{} transfer to real-world hardware?
\end{itemize}

\begin{table*}[t]
\centering
\caption{Comparison of our method to baselines in various terrains under the hard difficulty setting (i.e., flat push 0.6m/s, slope gradient 0.2, stair height 0.08m). The second row shows the road widths in meters. The best results are highlighted in boldface. Our method significantly outperforms the baseline methods.}
\label{tab:main}
\resizebox{\textwidth}{!}{
\begin{tabular}{l l ccc ccc ccc}
\toprule
\multirow{2}{*}{\textbf{Metric}} & \multirow{2}{*}{\textbf{Method}} & \multicolumn{3}{c}{\textbf{Flat}} & \multicolumn{3}{c}{\textbf{Slope}} & \multicolumn{3}{c}{\textbf{Stairs}} \\ 
\cmidrule(lr){3-5} \cmidrule(lr){6-8} \cmidrule(lr){9-11}
 &  & 0.25 & 0.3 & 0.35 & 0.25 & 0.3 & 0.35 & 0.25 & 0.3 & 0.35 \\ \midrule
\multirow{3}{*}{Success Rate} & {\ours{}} &  \textbf{0.90} $\pm$ 0.02 & \textbf{0.94} $\pm$ 0.02 & \textbf{0.99} $\pm$ 0.01 & \textbf{0.92} $\pm$ 0.02 & \textbf{0.96} $\pm$ 0.02 & \textbf{1.00} $\pm$ 0.00 & \textbf{0.89} $\pm$ 0.02 & \textbf{0.93} $\pm$ 0.02 & \textbf{0.98} $\pm$ 0.01  \\ 
& {\ours{} w/o Upper} & 0.00 $\pm$ 0.00 & 0.04 $\pm$ 0.03 & 0.18 $\pm$ 0.03 & 0.01 $\pm$ 0.01 & 0.13 $\pm$ 0.05 & 0.24 $\pm$ 0.04 & 0.00 $\pm$ 0.00 & 0.12 $\pm$ 0.04 & 0.25 $\pm$ 0.04  \\  
& {URG} & 0.01 $\pm$ 0.01 & 0.14 $\pm$ 0.12 & 0.12 $\pm$ 0.04 & 0.03 $\pm$ 0.02 & 0.18 $\pm$ 0.12 & 0.34 $\pm$ 0.21 & 0.02 $\pm$ 0.02 & 0.12 $\pm$ 0.08 & 0.22 $\pm$ 0.14  \\ \midrule
\multirow{3}{*}{MXD}  & {\ours{}} & \textbf{6.58} $\pm$ 0.17 & \textbf{6.78} $\pm$ 0.21 & \textbf{7.02} $\pm$ 0.20 & \textbf{5.83} $\pm$ 0.11 & \textbf{6.20} $\pm$ 0.15 & \textbf{6.52} $\pm$ 0.20 & \textbf{5.62} $\pm$ 0.20 & \textbf{5.79} $\pm$ 0.20 & \textbf{6.03} $\pm$ 0.23  \\ 
 &  {\ours{} w/o Upper} & 0.85 $\pm$ 0.04 & 1.16 $\pm$ 0.17 & 1.92 $\pm$ 0.25 & 1.07 $\pm$ 0.08 & 1.47 $\pm$ 0.15 & 1.81 $\pm$ 0.17 & 1.07 $\pm$ 0.06 & 1.55 $\pm$ 0.16 & 2.10 $\pm$ 0.16  \\ 
 &  {URG} & 1.33 $\pm$ 0.11 & 1.93 $\pm$ 0.73 & 2.02 $\pm$ 0.37 & 1.61 $\pm$ 0.24 & 2.24 $\pm$ 0.59 & 3.09 $\pm$ 1.04 & 1.48 $\pm$ 0.15 & 1.91 $\pm$ 0.43 & 2.42 $\pm$ 0.67  \\ \bottomrule
\end{tabular}
}
\vspace{-1em}
\end{table*}

\textbf{Experiments Setup.} 
We conduct experiments on Unitree H1-2, which is a full-sized humanoid robot with 27 DoF. The policy controls 21 DoF, excluding the 3 DoF in each wrist of the hands.
We train the RL policy in Isaac Gym \cite{makoviychuk2021isaac} and use MuJoCo \cite{todorov2012mujoco} as a \emph{sim-to-sim} verification platform. As shown in Fig.~\ref{fig:terr}, we visualize the different training terrains in the Isaac Gym simulation. In our experiments, we evaluate the performance across the three types of terrains (i.e., narrow flat, narrow slope, and narrow stairs), where each terrain has varying road widths and difficulties (i.e., 
maximum linear velocity of random push in flat, slope gradient in slope, and step height in stairs). For evaluation, we sample $10^4$ episodes with 20s duration and 0.5m/s heading velocity in Isaac Gym, and 100 episodes with 25s duration and 0.3m/s heading velocity in MuJoCo. The discrepancy in evaluation episode is due to MuJoCo is renowned for its high-fidelity physics modeling, enabling reliable results with fewer episodes. Meanwhile, the policy is trained on Isaac Gym and transferred to MuJoCo, which often leads to decreased performance. Thus, we adopt a longer duration and more conservative command for MuJoCo evaluation. We report the success rate (where success means the move distance $\geq 4\text{m}$) and the Mean X-Displacement (MXD) as metrics. 
    
\subsection{Result Comparison in Narrow Terrains}

\begin{figure}[t]
    \centering 
    \includegraphics[width=0.44\textwidth]{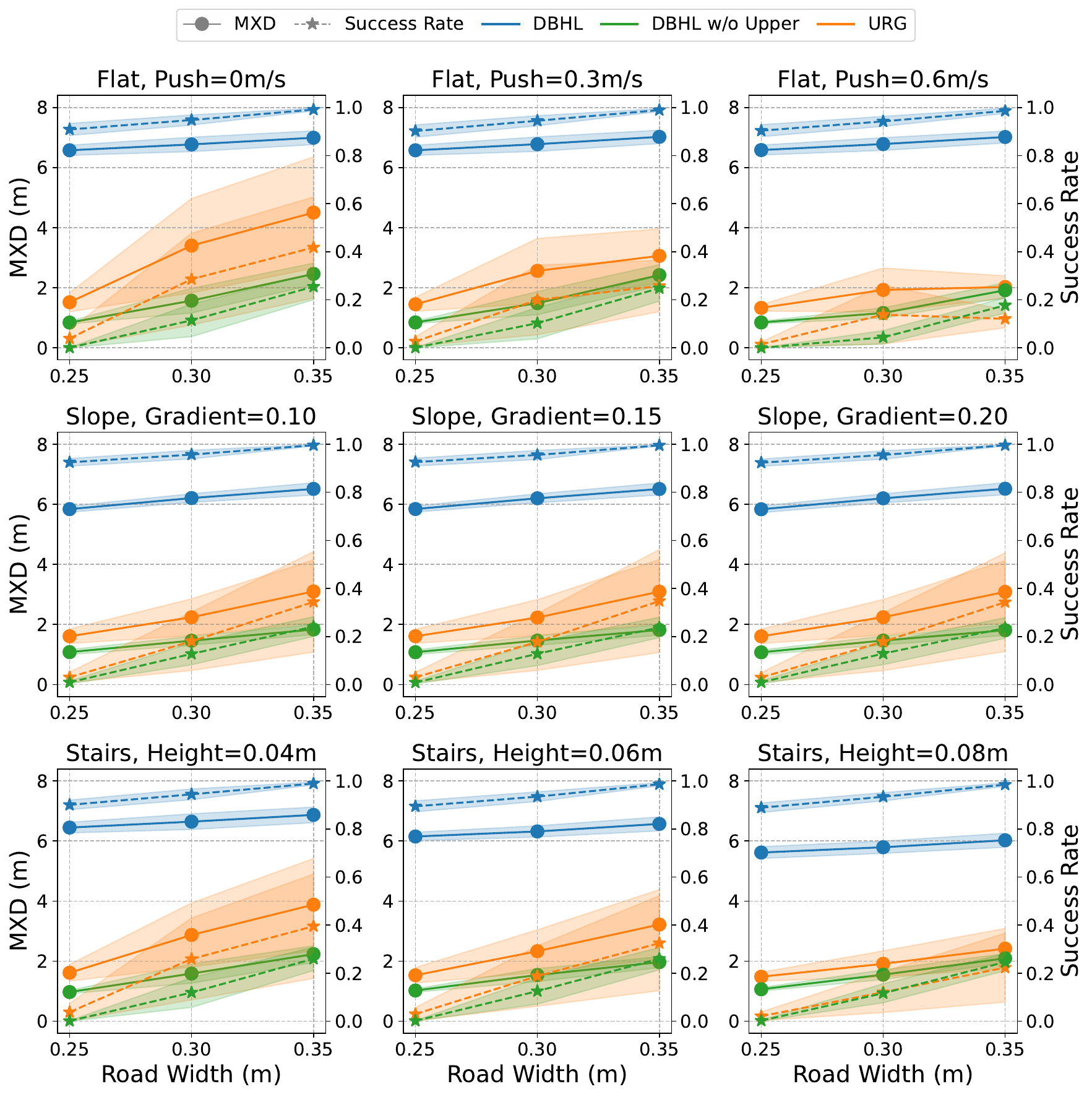}
    \caption{Comparison of our method to baselines in various terrains and difficulties. The result shows that whole-body control is essential for DBHL, and the dynamic balance mechanism is more effective than phase-based gait in challenging conditions. Each setting is evaluated over 3 random seeds. The shaded region around each curve represents $\pm1\sigma$ range, indicating the variability of the results.}
    \label{fig:main}
\vspace{-1em}
\end{figure}



To address Q1, we compare our methods with the following baselines in Isaac Gym:
\begin{itemize}
    \item \ours{} w/o Upper: This variant of our method fixes the upper body, focusing solely on the lower-body control. 
    \item Unitree RL Gym (URG): The official RL framework provided by Unitree, 
    which basically follows \cite{HumanGym1} by employing asymmetric actor-critic for lower-body control and relying on phase-based rewards to learn periodic gaits. We also incorporate the height field as part of the observation space to ensure a fair comparison. 
\end{itemize}

According to Fig.~\ref{fig:main} and Table~\ref{tab:main}, our method consistently outperforms the baselines in terms of both success rate and MXD. The results highlight the critical role of whole-body control in our approach. The performance drops significantly when only the lower body is controlled, indicating that narrow-terrain locomotion requires the coordinated effort of the entire robot. Meanwhile, our method demonstrates superior robustness compared to the phase-based control method across various terrains, achieving higher success rates and better MXD at all difficulty levels. This underscores the importance of phase-free locomotion and whole-body control in complex environments.

\subsection{{Analysis of ZMP-based Reward}}

To address Q2, we compare the performance of policy trained with and without the ZMP-based reward, focusing on dynamic balance and task performance.

\subsubsection{Quantitative Comparison}
\begin{figure}[h!]
    \vspace{-1em}
  \centering
  \includegraphics[width=0.45\textwidth]{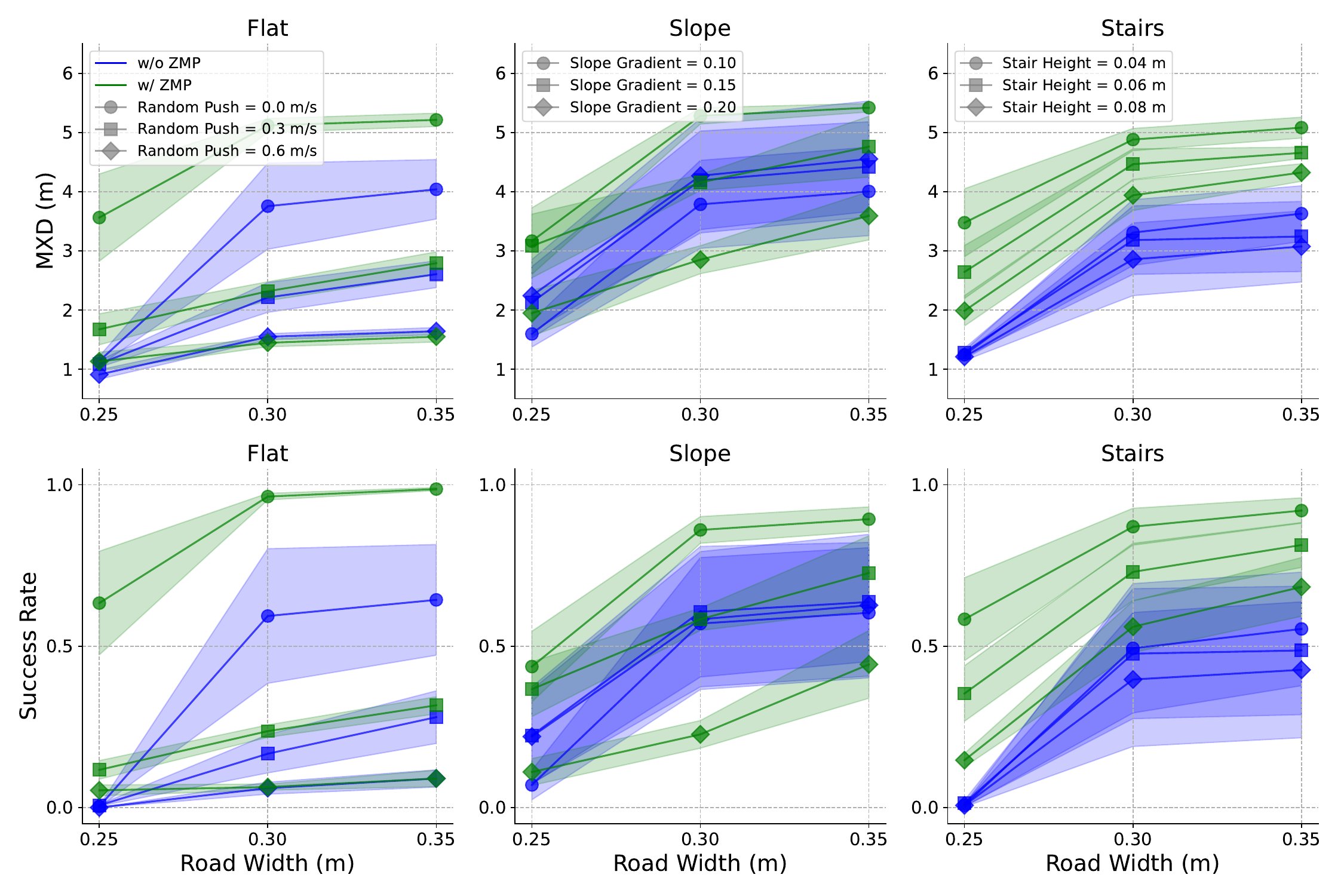}
  \caption{Quantitative comparison for ZMP-based reward. The inclusion of the ZMP reward improves the robot's dynamic stability, leading to a better success rate and MXD on narrow terrains. Each setting is evaluated over 3 random seeds. The shaded region around each curve represents $\pm0.5 \sigma$ range. }
  \label{fig:zmp_exp}
\end{figure}

Fig.~\ref{fig:zmp_exp} presents a quantitative comparison of success rate and MXD between variants with and without ZMP-based reward in MuJoCo. The reason to compare in MuJoCo is to leverage its high-fidelity simulation for sensitive detection of subtle policy differences. The results clearly show that incorporating ZMP-based rewards allows the robot to traverse narrow and challenging terrains more effectively than the non-ZMP variant, resulting in a significantly better success rate and MXD in most settings. An special case for non-ZMP variant arises when the slope gradient is 0.2, where the performance surpasses that observed on slopes of 0.1 and 0.15. We hypothesize that the non-ZMP variant may overfit to this condition, resulting in better performance compared to less challenging settings.

\subsubsection{Qualitative Comparison}
\begin{figure}[t]
  \centering
  
  \begin{subfigure}[c]{0.23\textwidth}
    \centering
    \includegraphics[width=\textwidth]{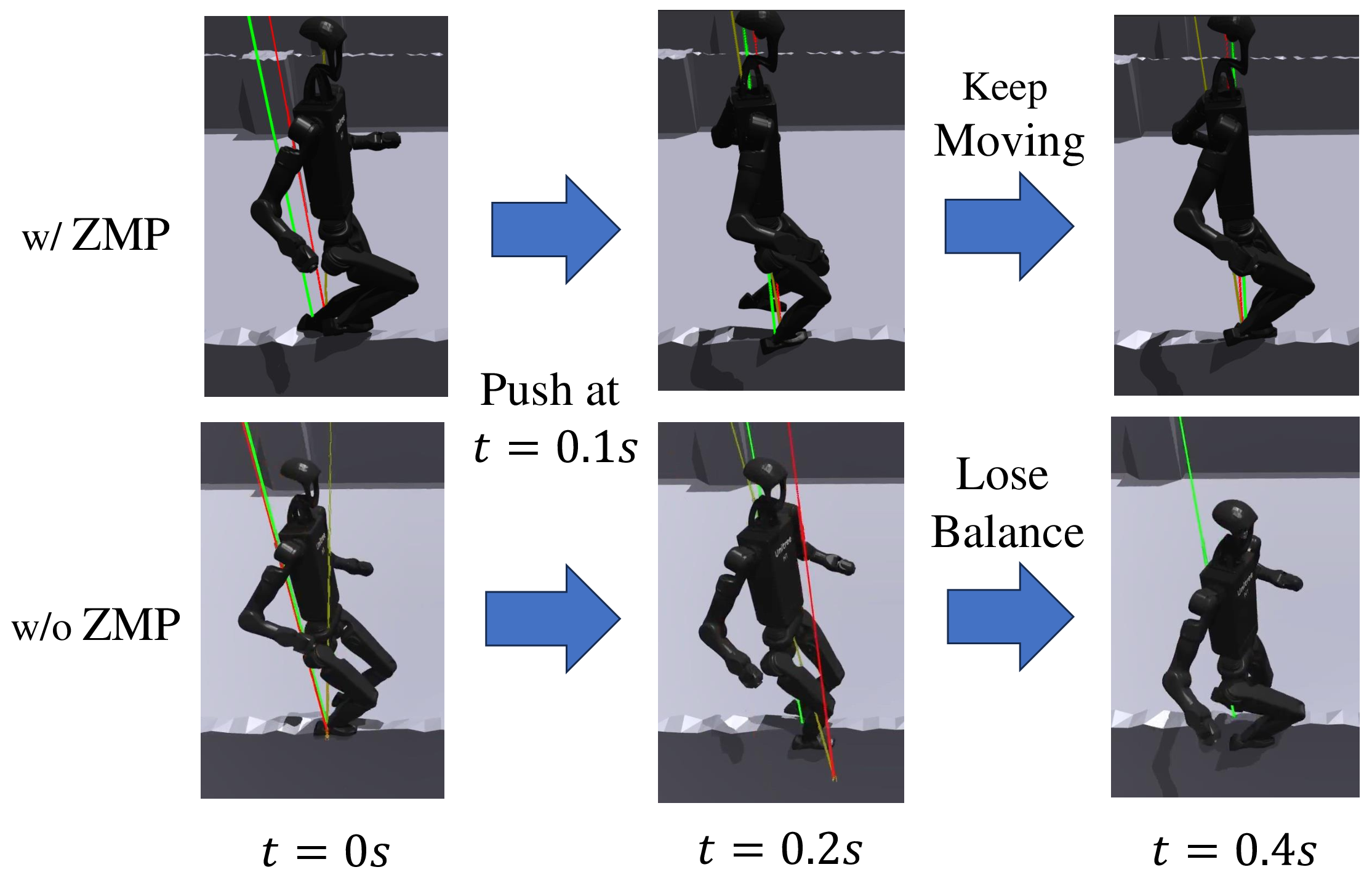}
    \caption{\small Snapshots of trajectories w/ and w/o. ZMP-based reward.}
  \end{subfigure}
  \hfill
  \begin{subfigure}[c]{0.23\textwidth}
    \centering
    \includegraphics[width=\textwidth]{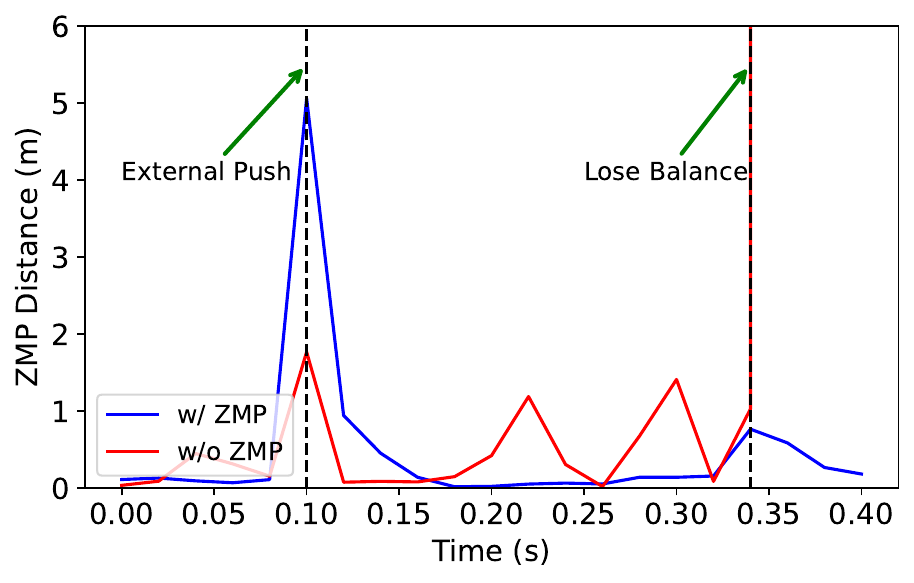}
    \caption{\small ZMP-distance over time.}
  \end{subfigure}
  \caption{Comparison of two example trajectories under external pushes. Without the ZMP-based reward, the robot loses balance and falls after pushing, as observed by the diverging ZMP-distance. In contrast, \ours{} with ZMP rewards maintains dynamic balance. In snapshots, the yellow lines represent ZML while the green and red lines represent vertical lines crossing $\bm p_\text{csp}$ and $\text{Proj}_\text{ZMP}(\bm p_\text{csp})$, respectively.}
  \label{fig:zmp_sample}
\end{figure}

In Fig.~\ref{fig:zmp_sample}(a), we provide two example trajectories for a qualitative comparison. At $t=0.1$s, an external push ($\Delta v_y=0.6$m/s) is exerted on the robot. The trajectory without ZMP-based reward results in the robot losing balance, where the corresponding ZMP-distance, defined as $\|\bm{p}_\text{csp} - \text{Proj}_\text{ZML}(\bm{p}_{\text{csp}})\|_2$ in Eq.~\eqref{eq:rzmp}, becomes particularly large and eventually diverges, as shown in Fig.~\ref{fig:zmp_sample}(b). In contrast, the policy incorporating ZMP-based reward successfully maintains the robot's stability throughout the disturbance since the distance is constrained by maximizing the ZMP-reward in Eq.~\eqref{eq:rzmp}.

\subsection{Ablation Study}

To address Q3, we perform ablation studies in Isaac Gym to investigate the effect of two key design choices: reward vectorization and multiplicative action noise.

\subsubsection{Effect of Reward Vectorization}

\begin{figure}[t]
  \vspace{-0.5em}
  \centering
  \includegraphics[width=0.38\textwidth]{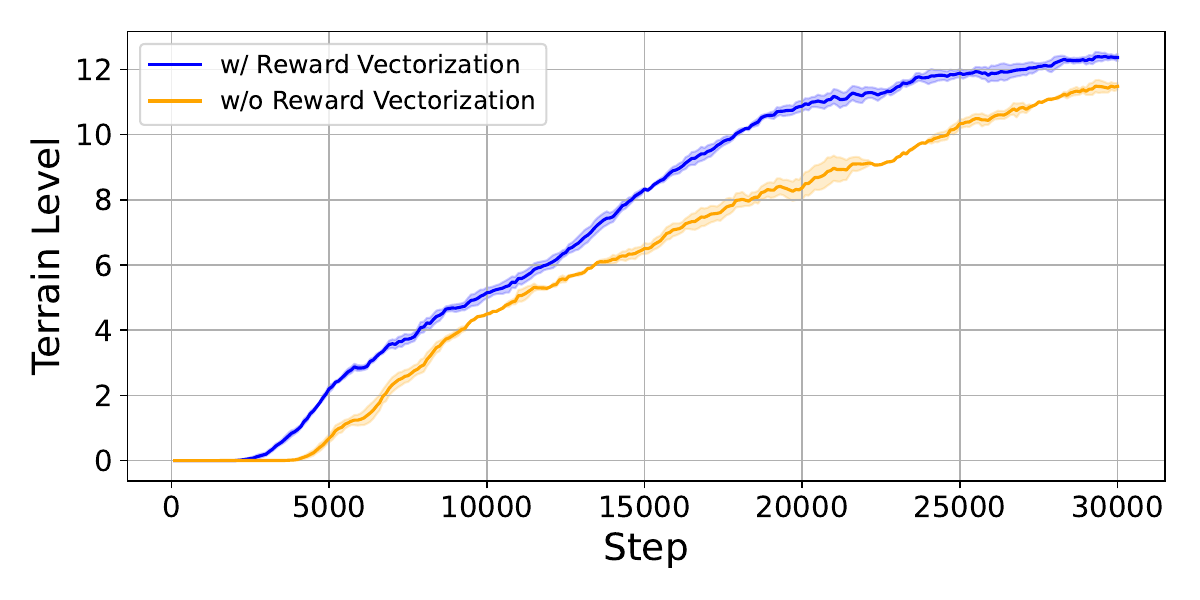}
  \vspace{-1em}
  \caption{Comparison of training curves with and without reward vectorization. 
  The terrain level represents the average difficulty level determined by the terrain-curriculum mechanism, with a total of 20 predefined difficulty levels. Each setting is evaluated over 3 random seeds. The training curves are sampled at 100-step intervals and smoothed using an exponential moving average with a smoothing factor of 0.1. }
  \label{fig:abl_rv}
  \vspace{-1em}
\end{figure}

Fig.~\ref{fig:abl_rv} shows the training curves for the policy with and without reward vectorization. The use of reward vectorization significantly accelerates the learning process, facilitating faster policy convergence. This improvement is attributed to the reward and value vectorization framework, which enables DBHL to learn each value term associated with each reward term independently, thus increasing overall learning efficiency.

\subsubsection{Effect of Multiplicative Action Noise}
\begin{figure}[t]
  \centering
  \begin{subfigure}[c]{0.32\textwidth}
      \centering
      \includegraphics[width=\textwidth]{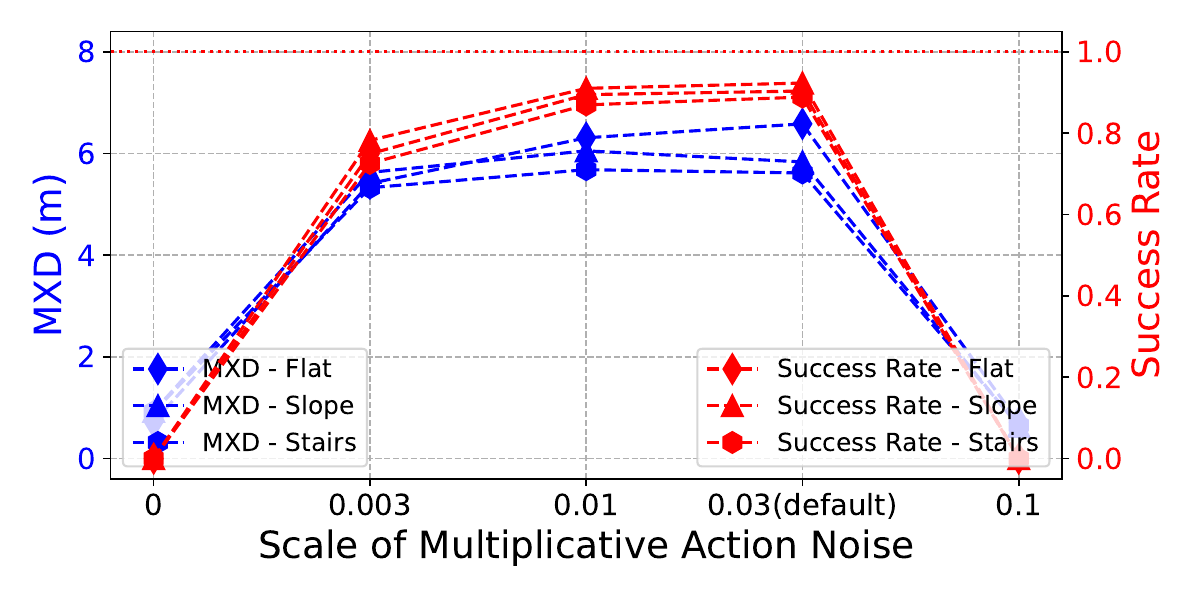}
      \caption{\small Success rate and MXD for varying scales of the action noise.}
  \end{subfigure}
  \hspace{0.5em}
  \begin{subfigure}[c]{0.32\textwidth}
      \centering
      \includegraphics[width=\textwidth]{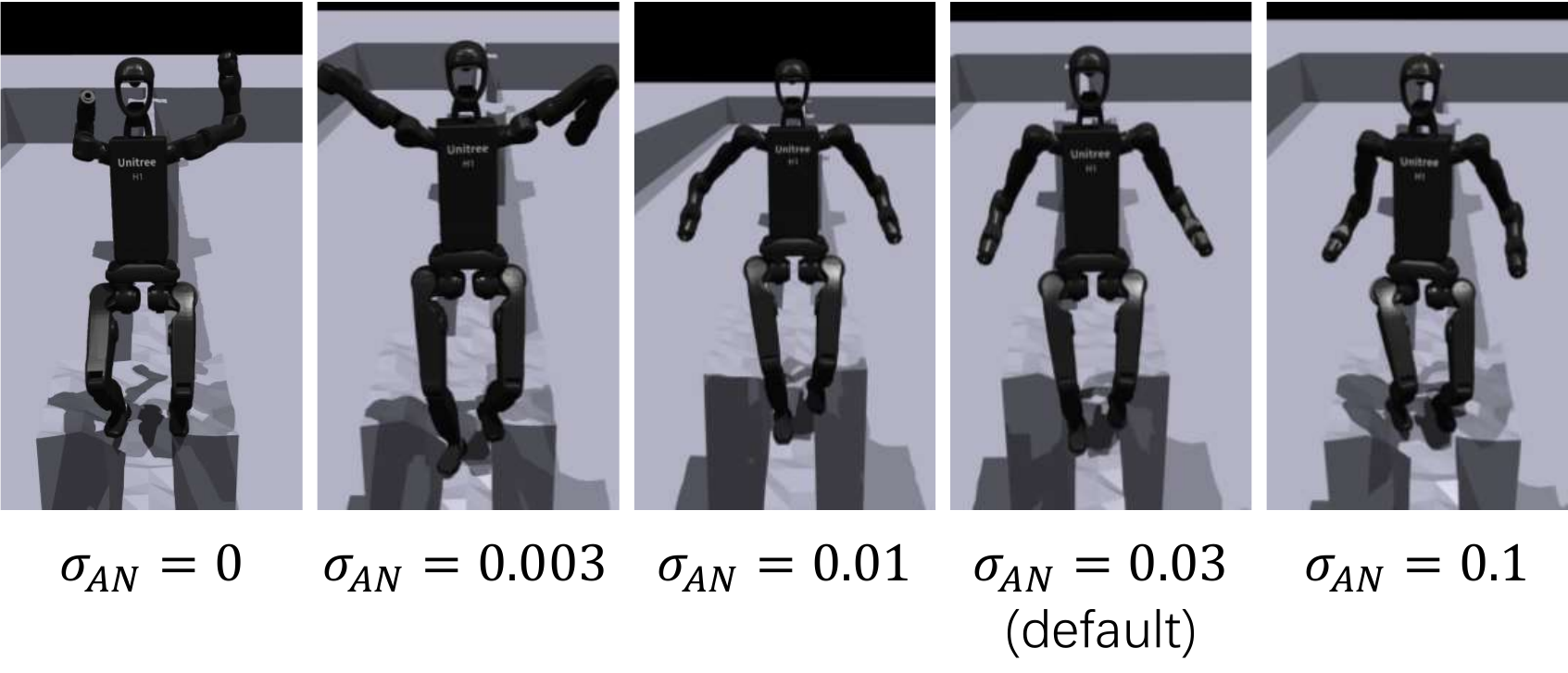}
      \caption{Snapshots of the policy behaviors for varying scales of the action noise.}
  \end{subfigure}
  \caption{Ablation study for action noise. A proper scale of action noise can constrain the range of motion of the upper body joints and improve task performance. We conduct experiments on narrow terrains with a width of 25cm under the hard difficulty setting. }
  \label{fig:abl_an}
  \vspace{-0.5em}
\end{figure}

Fig.~\ref{fig:abl_an} presents the results for varying scales of action noise. Our experiments indicate that an appropriately calibrated level of action noise effectively confines the motion range of the upper body joints. Conversely, both excessively low and high levels of action noise lead to instabilities in accomplishing the task.

\begin{figure}[t]
  \centering
  \includegraphics[width=0.5\textwidth]{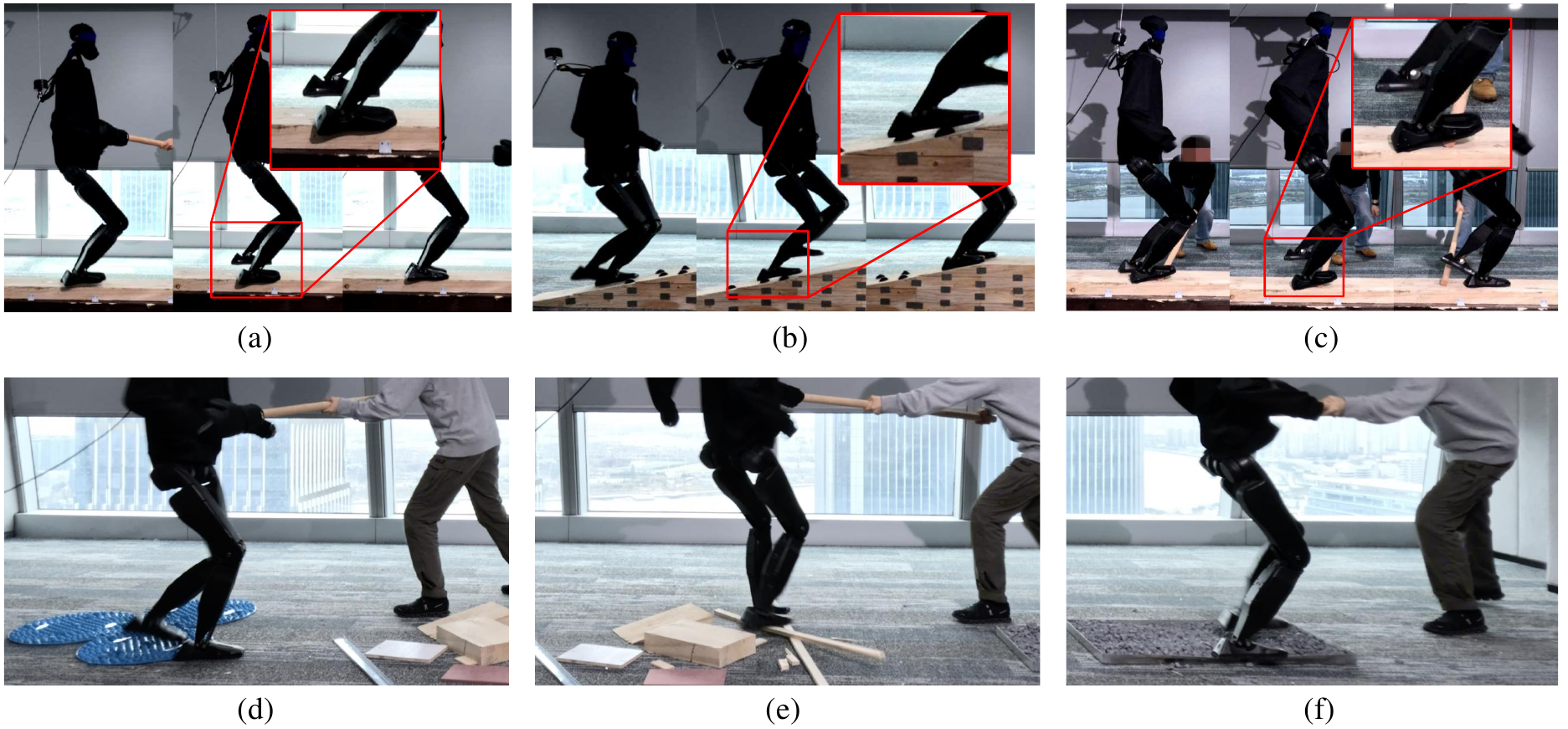}
  \caption{Real-world experiments, including (a) narrow path under perturbations, (b) sloped terrain with conical obstacles, (c) moving stick trips, (d) acupressure plates, (e) wooden block obstacles, and (f) stone roads, showcasing DBHL's efficacy in addressing real-world challenges.}
  \label{fig:real_beam}
  \vspace{-1em}
\end{figure}

\subsection{Real-World Experiment}

To address Q4, we deploy \ours{} on the Unitree H1-2 robot and evaluate its performance in both narrow and planar terrains through a series of real-world experiments. The evaluation platform consists of a narrow wooden track measuring 25 cm in width. This track includes three sections: a sloped ramp with a gradient of 0.2 and a length of 1.6 m, a bridge section spanning 1.6 m, and a set of stairs with a step width of 40 cm and a step height of 8 cm.

\subsubsection{External Disturbances in Narrow Terrains}
 As shown in Fig.~\ref{fig:demo} and Fig.~\ref{fig:real_beam} (a-c), \ours{} can traverse the narrow track successfully. The controller effectively handles a range of challenging disturbance conditions, including carrying a 5kg payload, passing through dense conical obstacles, withstanding human pushes, and walking on a levered stick. 

\subsubsection{Irregular Obstacles in Uneven Ground} Since the previous experiments focus solely on narrow terrains, we train an additional policy on irregular planar terrain. Then we evaluate the policy's capabilities by incorporating a variety of irregular obstacles on flat terrain in the real world, as depicted in Fig.~\ref{fig:real_beam} (d-f). This environment features acupressure plates, wooden blocks, planks, and stone-paved paths. In addition, the robot must contend with applied external forces while traversing challenging conditions. The results show DBHL can overcome these complex terrain challenges.


\section{CONCLUSIONS}

This work presents DBHL, a novel reinforcement learning framework that enables humanoid robots to traverse extreme terrains by introducing a ZMP-based reward function and a whole-body control framework. Through extensive simulations and real-world experiments, DBHL demonstrates superior performance in narrow and uneven terrains, highlighting its robustness and adaptability. 
The proposed methodology opens new possibilities for real-world applications requiring extreme mobility and balance.




\section*{ACKNOWLEDGMENT}

This work is supported by the National Natural Science Foundation of China (Grant No.62306242).



\bibliographystyle{IEEEtran}
\bibliography{ref}

\end{document}